%% file: sample-sigconf-authordraft.tex
\documentclass[sigconf]{acmart}

\AtBeginDocument{%
  }

\usepackage{algorithm}
\usepackage{algpseudocode}

\usepackage{amsmath,amsfonts}
\usepackage{graphicx}
\usepackage{textcomp}
\usepackage{xcolor}
\usepackage{booktabs} % for tables
\usepackage{caption} % for subfigures
\usepackage{subcaption} % for subfigures

\copyrightyear{2025}
\acmYear{2025}
\setcopyright{rightsretained}
\acmConference[FDG '25]{International Conference on the Foundations of Digital Games}{April 15--18, 2025}{Graz, Austria}
\acmBooktitle{International Conference on the Foundations of Digital Games (FDG '25), April 15--18, 2025, Graz, Austria}
\acmPrice{}
\acmDOI{10.1145/3723498.3723747       }
\acmISBN{/25/04}

\begin{document}

\title{Level the Level: Balancing Game Levels for Asymmetric Player Archetypes With Reinforcement Learning}

\author{Florian Rupp}
\email{f.rupp@hs-mannheim.de}
%\email{webmaster@marysville-ohio.com}
\affiliation{%
  \institution{Mannheim Technical University}
  \city{Mannheim}
  %\state{State}
  \country{Germany}
}

\author{Kai Eckert}
\email{k.eckert@hs-mannheim.de}
%\email{webmaster@marysville-ohio.com}
\affiliation{%
  \institution{Mannheim Technical University}
  \city{Mannheim}
  %\state{State}
  \country{Germany}
}

\renewcommand{\shortauthors}{Rupp et al.}
\renewcommand{\shorttitle}{Level the Level: Balancing Game Levels for Asymmetric Player Archetypes with RL}

\include{content/abstract}





\maketitle

\input{content/content}

\bibliographystyle{ACM-Reference-Format}
\bibliography{bibliography}

%\printbibliography

\end{document}

%% file: content/abstract.tex
\begin{abstract}
Balancing games, especially those with asymmetric multiplayer content, requires significant manual effort and extensive human playtesting during development. For this reason, this work focuses on generating balanced levels tailored to asymmetric player archetypes, where the disparity in abilities is balanced entirely through the level design. For instance, while one archetype may have an advantage over another, both should have an equal chance of winning.
We therefore conceptualize game balancing as a procedural content generation problem and build on and extend a recently introduced method that uses reinforcement learning to balance tile-based game levels.
We evaluate the method on four different player archetypes and demonstrate its ability to balance a larger proportion of levels compared to two baseline approaches. Furthermore, our results indicate that as the disparity between player archetypes increases, the required number of training steps grows, while the model's accuracy in achieving balance decreases.
\end{abstract}

%% file: content/content.tex
\section{Introduction}

Game levels for competitive play need to be balanced in order to ensure player satisfaction and to avoid frustration or boredom~\cite{becker_what_2020}. The process of balancing a game, however, involves a lot of manual work, effort, and human play testing~\cite{schreiber_game_2021}. For this reason, many works propose approaches to automate the process of game balancing~\cite{beau_automated_2016,volz_demonstrating_2016,pfau_dungeons_2020,morosan_automated_2017,lanzi_evolving_2014,lara-cabrera_balance_2014,rupp_geevo_2024}.
Well-designed games ensure the viability of multiple strategies for players to choose from, all of which, if played well, can lead to victory~\cite{schreiber_game_2021}.
As a result, many modern games make use of asymmetric balancing strategies, such as putting heroes of different abilities against each other.
%This can be seen in dungeon crawlers like \emph{Decent: Journeys in the Dark} or MOBA (Multiplayer Online Battle Arena) games like \emph{League of Legends}. 
The asymmetric balance is mainly achieved by balancing the numerical values of the game units, such as health or attack values.

Limited work~\cite{beau_automated_2016,lanzi_evolving_2014,rupp_geevo_2024}, however, has been published to automatically balance asymmetric games, i.e. players with different abilities and stats.
For this reason, this work focuses on balancing an asymmetric game setup on the example of players with different abilities entirely through \emph{level design}, e.g., where to place resources in relation to the players' spawn positions, formulating this task as a procedural content generation (PCG) problem. 
This approach is further motivated to be used in settings where players of different skill levels face each other, such as experts and beginners, or adults playing against children. Another use case is to ensure balance in competitive settings where players with different gear levels have different strengths.

Recently, the Procedural Content Generation via Reinforcement Learning framework (PCGRL) has shown promising results in creating tile based levels~\cite{khalifa_pcgrl_2020}. It has been extended to apply to game balancing~\cite{rupp_balancing_2023,rupp_simulation_2024}, where the balance of a level is determined through multiple simulations using heuristic agents. A level is considered balanced when all players win equally often.

This approach is, however, limited by the fact that it only uses exactly the same archetype of agents playing against each other. To address this shortcoming, this work adds four new archetypes to investigate the ability of the method to achieve balance for \emph{different} agent archetypes playing against each other, for instance, a stronger agent playing against a handicapped agent.
We evaluate and compare our results to a random search and a hill-climbing baseline. In addition, we improve the action space introduced in~\cite{rupp_balancing_2023} by halving the size of the action space to speed up convergence in training.
The code of the introduced heuristic agents and the game environment can be found on Github.\footnote{https://github.com/FlorianRupp/feast-and-forage-env} Our contributions are:
\begin{itemize}
    \item The application of RL to balance game levels for several asymmetrically paired heuristic player archetypes entirely through the level design.
    \item An improvement in the definition of the action space for the Markov Decision process that halves the size of the action space resulting in faster convergence in training.
    \item A study to evaluate the results against a random search and a hill-climbing baseline.
    %\item A re implementation of the game environment used in~\cite{rupp_balancing_2023} to speed up simulations and additional heuristic agents.
\end{itemize}

\renewcommand{\thefootnote}{}
\footnote{This research was supported by the Volkswagen Foundation (Project:
Consequences of Artificial Intelligence on Urban Societies, Grant 98555)}

%The asymmetric balancing of game levels with RL on the example of several different paired-up heuristic player archetypes entirely through level design.

\section{Related Work}
There are several approaches to balancing a game, one of which is to configure the values of the game entities~\cite{schreiber_game_2021}.
For asymmetric games, Beau et al.~\cite{beau_automated_2016} propose an approach using Monte Carlo simulations to estimate the effects of game actions, and demonstrate how to find imbalances in the value configuration. Like our approach, they also optimize imbalances until they are balanced according to the simulation approximations. Evolutionary algorithms have been used by Volz et al.~\cite{volz_demonstrating_2016} to balance and build decks for card games, by De Mesentier Silva~\cite{mesentier_silva_evolving_2019} to balance the meta of a collectible card game, by Morosan et al.~\cite{morosan_automated_2017} for real-time strategy (RTS) games, and for game economies by Rupp et al.~\cite{rupp_geevo_2024}. Pfau et al.~\cite{pfau_dungeons_2020} optimize game parameters using data-driven deep player behavior models.

In contrast, this work focuses on the automated balancing of games through the level design entirely using PCG. Lanzi et al.~\cite{lanzi_evolving_2014} introduced an evolutionary algorithm for balancing maps for first-person shooters. The authors also include the balancing of asymmetries by using bots with different weapon types. Lara Cabrera et al.~\cite{lara-cabrera_balance_2014} balance game maps for an RTS game, also using an evolutionary algorithm.
This work builds on and extends the works of Rupp et al.~\cite{rupp_balancing_2023,rupp_simulation_2024}, which introduce RL to balance tile-based levels. A more detailed explanation for better understanding is given in the Background chapter (Section~\ref{sec:pcg-balancing}).

%In addition to game balancing, PCG is also an active field of research to generate game content such as game levels in general~\cite{togelius_search-based_2011,khalifa_pcgrl_2020}, rules~\cite{cook_mechanic_2013}, quests~\cite{ashby_personalized_2023}, or narratives~\cite{alvarez_tropetwist_2022}.

\section{Background}

\subsection{Game Environment}
The game (see Figure~\ref{fig:samples}) is a survival challenge for two players on a 6x6 tile-based map in the Neural Massively Multiplayer Online (NMMO)
environment~\cite{suarez_neural_2019}.
To win, players must either collect a specific amount of food resources or survive longer than their opponent.
There are four tile types: grass~\raisebox{-0.2em}{\includegraphics[width=1em]{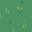}}, rock~\raisebox{-0.2em}{\includegraphics[width=1em]{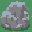}}, water~\raisebox{-0.2em}{\includegraphics[width=1em]{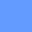}}, and food~\raisebox{-0.2em}{\includegraphics[width=1em]{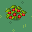}}. Movement is hindered by rock and water tiles.
%Win conditions or the impact of tiles are specifically changed for particular archetypes as we will introduce in Section~\ref{sec:a-balancing}.
Simultaneously, players choose from one of the five actions: up, down, left, right or to do nothing. Their state consists of position, health, and water and food levels.
Each turn, players lose water, food, and also health if both water and food reach zero.
Players lose when their health reaches zero. Players can refill food by moving on food tiles, which then become scrub tiles. Scrub tiles can respawn as food tiles with a chance of 2.5\% each turn.
Water can be refilled by moving onto tiles adjacent to water tiles, which are never depleted. If food and water levels exceed 50\%, health is gradually restored.
The red player~\raisebox{-0.2em}{\includegraphics[width=1em]{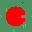}} is player one and the yellow player~\raisebox{-0.2em}{\includegraphics[width=1em]{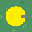}} is player two. In this work, both players are controlled by heuristic agents based on a greedy strategy, always collecting the nearest available food resource (Algorithm~\ref{alg:forage-agent}). For the experiments, this heuristic will be slightly modified to create several different heuristic archetypes (cf. Section~\ref{sec:a-balancing}).

\subsection{PCGRL for Game Level Balancing}
\label{sec:pcg-balancing}
Khalifa et al. introduced the PCGRL framework~\cite{khalifa_pcgrl_2020} to model level generation as a Markov decision process (MDP), leveraging reinforcement learning (RL). In a trajectory of multiple actions, the RL agent modifies the level until it meets predefined constraints expressed by the reward function.

Based on this approach, Rupp et al. introduced an approach using PCGRL to balance previously generated levels~\cite{rupp_balancing_2023,rupp_simulation_2024}. Balancing is considered as a fine-tuning process for existing levels, which has been shown to converge faster as when levels are generated and balanced in a single step~\cite{rupp_balancing_2023}.
The action space is predicting which two tiles to swap. A level's balance is determined by simulating the game multiple times with two identical heuristic agents. The reward then is based on the balance, determined by the frequency of each agent's victory in the game.
%and can be understood as a static simulation-based evaluation function. 
%as defined in~\cite{yannakakis_experience-driven_2011}.
The reward function evaluates a level's state by assigning a value between 0 and 1, where 0.5 signifies equal win rates for both players, and 0 or 1 indicates that one player wins every round.
An additional study with human play testers concludes that the balancing simulated with artificial heuristic agents actually improves the perceived balance for humans in most cases~\cite{rupp_might_balanced_2024}.

Unlike search-based approaches, RL can generate content fast once the model has been trained. This is particularly important here as the simulations used in the reward step are computationally expensive. As RL learns during training, it learns to avoid unnecessary simulation steps, speeding up inference time. 
Other works extend PCGRL for controllability~\cite{earle_learning_2021}, 3D levels~\cite{jiang_learning_2022-4}, graph data~\cite{rupp_gpcgrl_2024}, and scalability~\cite{earle_scaling_2024}.
All PCGRL approaches use Proximal Policy Optimization (PPO)~\cite{schulman_proximal_2017} as RL algorithm.

\begin{algorithm}
\caption{Heuristic Agent Archetype A as used in~\cite{rupp_balancing_2023,rupp_simulation_2024}.}
\label{alg:forage-agent}
\begin{algorithmic}[1]

\Procedure{Step}{gameState}
    \State \textbf{init} \textit{action} $\leftarrow$ \textbf{DoNotMove}
    \State \textbf{init} \textit{foodReachable} $\leftarrow$ \Call{FoodReachable}{gameState}
    \State \textbf{init} \textit{waterReachable} $\leftarrow$ \Call{WaterReachable}{gameState}
    
    \If{\textit{foodReachable}}
        \State \textit{action} $\leftarrow$
        \Call{FindShortestPathTo}{food}
    \ElsIf{\textit{waterReachable}}
        \State \textit{action} $\leftarrow$
        \Call{FindShortestPathTo}{water}
    \EndIf
    \State \Return action
\EndProcedure
\end{algorithmic}
\end{algorithm}

\section{Method}

\subsection{Improving the Action Space}
\label{sec:action-space}

For this work, we use the \emph{swap-wide} representation to define the action space of the MDP, as it gives the best results in~\cite{rupp_simulation_2024}. This action space allows the RL agent to swap the tile of a predicted location $(x1,y1)$ with the tile of a second predicted location $(x2,y2)$ of a level. With another flag it can predict if the swap should be done or not. This results in an action space depending on the height $h$ and width $w$ of the level: $[h,w,h,w,2]$. For a 6x6 level this results in 2592 actions.
Since the model already makes a prediction about the positions where to swap, the additional prediction of whether to swap or not makes the action space unnecessarily complex. Therefore, we reduce it to predict only the two swap positions, resulting in the action space: $[h,w,h,w]$. For a 6x6 level this reduces the action space to the half of 1296 actions.

\subsection{Asymmetric Balancing}
\label{sec:a-balancing}

In this work, we introduce the four new agent archetypes B, C, D1, and D2, which extend the existing archetype A (Algorithm~\ref{alg:forage-agent}).
Agents are implemented by a \textsc{step} function, which is called once per turn for each agent. The step function always returns one of the five actions, describing which of the four adjacent tiles to move to next, or to do nothing.
The helper functions \textsc{FoodReachable} and \textsc{WaterReachable} use path-finding to determine whether there is a valid path from the agent's position to at least one of the respective resources.
If so, the A* algorithm is used to find the shortest path, always favoring food resources as they have a greater impact on winning the game.
Each new archetype addresses a specific aspect of the balance, such as movement advantages or gaining victory points by consuming food. 
The list below gives a brief description of the agents used and how this affects their chance of victory compared to archetype A:
%We define the following archetypes:
\begin{itemize}
    \item Archetype A is the \emph{Base Agent}, as described in Algorithm~\ref{alg:forage-agent} and applied in~\cite{rupp_balancing_2023,rupp_simulation_2024}. It cannot move over rock and water tiles and wins with five victory points.
    \item Archetype B, the \emph{Rock Agent}, has the additional ability to cross rock tiles, being blocked only by water tiles. This gives it an advantage over archetype A agents.
    \item Archetype C, the \emph{Handicap Agent}, can only perform one action every second turn. This agent is at a huge disadvantage when playing against archetype A agents.
    \item Archetype D, the \emph{Food Agent}, already wins the game with four (D1) or three (D2) collected food resources instead of five. This gives it an advantage over archetype A agents.
\end{itemize}

\section{Results, Discussion and Limitations}

For all experiments, we use the same dataset of 500 generated levels to ensure a fair comparison. We compare our PCGRL method with two baselines: a random agent and a simple hill-climbing approach as used in~\cite{rupp_simulation_2024}. The hill-climbing approach uses the same swapping mechanism as PCGRL, but chooses the positions randomly. If the reward is not positive, it transitions back to the previous state.
A key metric for evaluation is the proportion of how many levels a method can balance.

We train multiple models where archetype A faces the new ones in a 1v1 setting. For all PPO models we use two 3-layered multi-layer perceptrons with layers of sizes 64, 128, and 64 as feature extractor and the value function. Each model had to be trained for a different number of steps until convergence, for comparison 10 million steps would result in 326 policy updates. We use a step size of 512 and parallelize 60 environments. This results in 30,720 trajectories for updating the policy and neuronal networks per single update.

\subsection{Performance and Comparison to Baselines}

All models were trained with the new action space definition in Section~\ref{sec:action-space} and a model with the existing action space definition from~\cite{rupp_balancing_2023}. In a direct comparison, the training using the smaller action space converges faster for all models and allows for higher rewards, resulting in overall better performance, in particular for type A vs. A (cf. Table~\ref{tab:results} and also~\cite{rupp_balancing_2023,rupp_simulation_2024}).

\begin{figure}
\vspace{-1mm}
    \centering
    \includegraphics[width=0.8\linewidth]{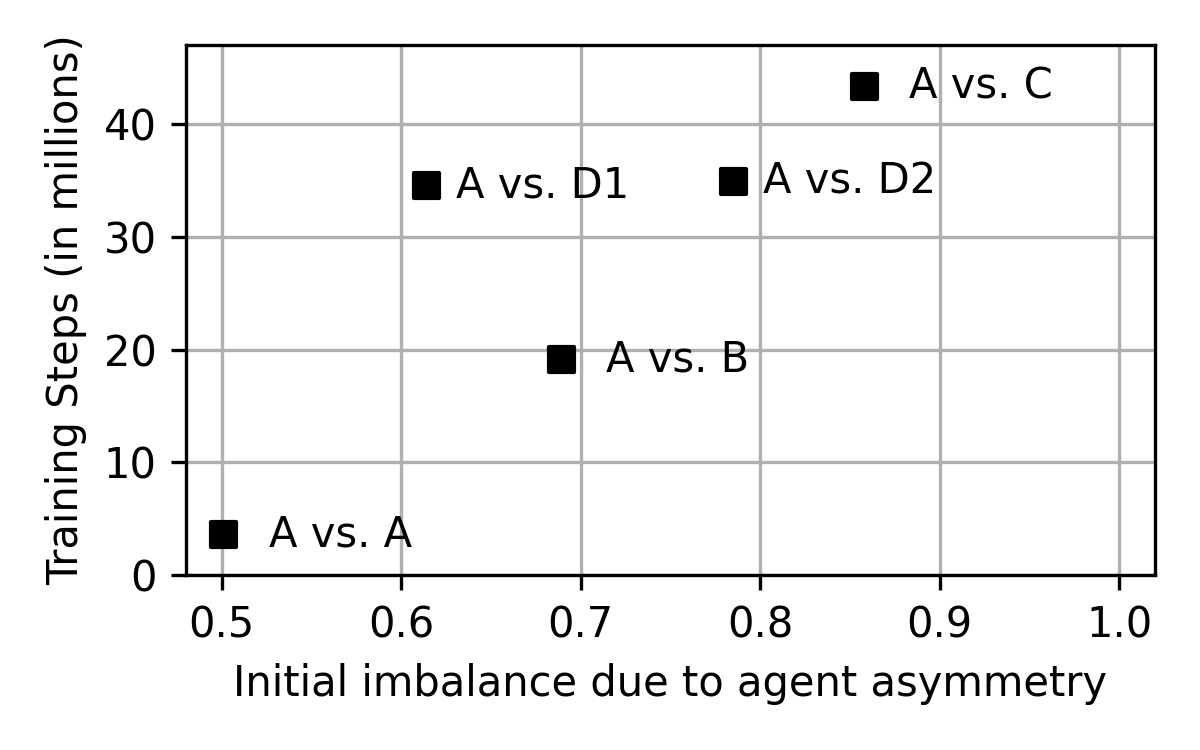}
  \caption{Training steps required for model convergence compared to the initial imbalance due to agent asymmetries per setup. A value of 0.5 indicates equal wins, while 1.0 means one agent always wins.}
    \label{fig:unfairness}
\end{figure}

To assess the imbalance of a particular archetype setup, we calculate the proportion of dataset levels initially biased towards that agent, where the agent wins more often in simulations than its opponent.
Figure~\ref{fig:unfairness} shows the relationship between initial imbalance and the number of training steps required for convergence.
%When inspecting the initial balance for the different agent settings in the level dataset, we see larger proportions balanced towards a specific player. 
When the archetypes setup is symmetric (A vs. A), the proportions of levels that initially favor a single player are almost equal (50.1\%). 
%In contrast, for the setup of A vs. B 68.9\%, A vs. C 85.8\%, and A vs. D1 61.4\%, A vs. D2 78.5\% are heavily unbalanced towards a specific player.
In contrast, the asymmetric setups strongly favor one specific player.
Comparing the initial imbalance with the number of training steps required for model convergence shows that the greater the disparity in strength between two archetypes of a setup, the more training steps are required.
This method is also useful for accurately measuring the disparity in asymmetry caused by differing abilities.
%From this we conclude that type B and D have an advantage over type A based on the agent's skills, and that type C has a strong disadvantage over type A. This is in line with our previous intention regarding the design of the agents.

Table~\ref{tab:results} shows the performance of the various setups compared to two baselines. Initially balanced levels are not taken into account. While the hill-climbing approach achieves reasonable results, our PCGRL approach remains the best in comparison for all setups. This is due to the advantage of RL to learn during training which trajectories have the best impact on the balance.

We also see that the performance of the different archetypes varies by about 20 percentage points. In relation to Figure~\ref{fig:unfairness} we observe that the greater the initial disparity between the two archetypes is, the more the performance of the models decreases. We can therefore conclude that the greater the initial unfairness of a setup, the harder it is for the model to learn how to compensate the balance by modifying the level alone.

% A vs. B 27\%, A vs. C 88\%, and A vs. D 19\% 

\begin{table}
    \centering
    \caption{Comparison of the proportions of balanced levels on a set of 500 generated levels with two baseline approaches. A level is balanced when both players win equally.}
    \begin{tabular}{ccccc} \toprule
           \textbf{Agents}&  \textbf{Random (\%)} & \textbf{Hill-Climbing (\%)} & \textbf{PCGRL (\%)} \\
    \cmidrule(lr){1-1} \cmidrule(lr){2-2} \cmidrule(lr){3-3} \cmidrule(lr){4-4} \cmidrule(lr){5-5}           
        A vs. B &  27.6 &  46.1  & \textbf{80.4} \\
        A vs. C &  16.2 &  24.6 & \textbf{56.5} \\
        A vs. D1 &  28.8 &  46.6 & \textbf{72.3} \\ 
        A vs. D2 &  26.8 &  35.2 & \textbf{57.9} \\ \midrule
        A vs. A~\cite{rupp_simulation_2024} & n.a. &  59.6 & 68.0 \\
        A vs. A &  55.8 &  59.6 & \textbf{89.7} \\
        \bottomrule
    \end{tabular}
    \label{tab:results}
\end{table}

\begin{figure}[htbp] % Use figure* for spanning across columns
    \centering
    % Main subfigure 1
    \begin{subfigure}[t]{0.24\textwidth} % Half of total width
        \centering
        % Nested subfigure 1.1
        \begin{subfigure}[t]{0.5\textwidth} % 50% of the parent
            \centering
            \includegraphics[width=\textwidth]{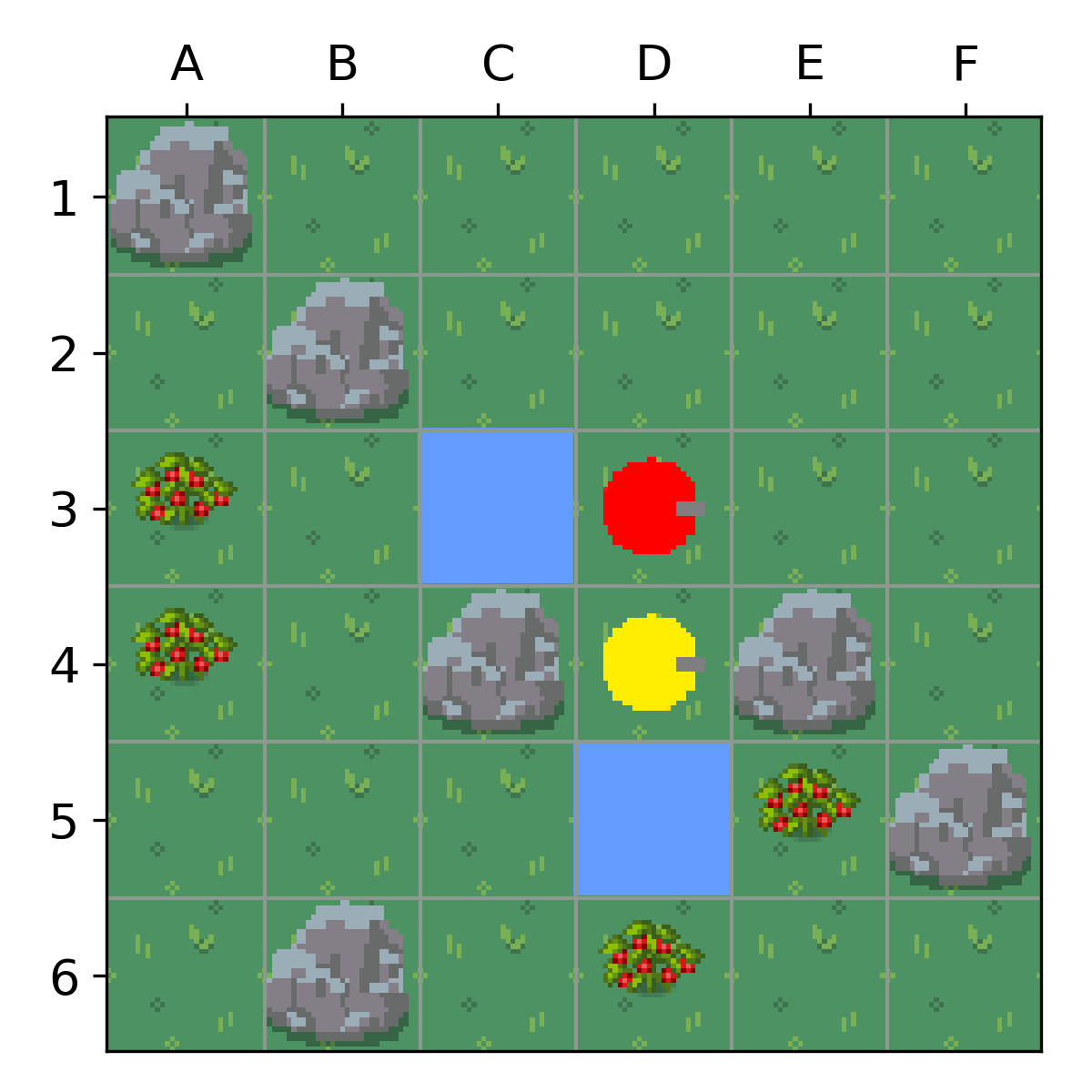}
            %\caption*{Before 1.0}
            \centering \footnotesize Unbalanced, 1.0
            \label{fig:subsubfig1}
        \end{subfigure}%
        \begin{subfigure}[t]{0.5\textwidth} % 50% of the parent
            \centering
            \includegraphics[width=\textwidth]{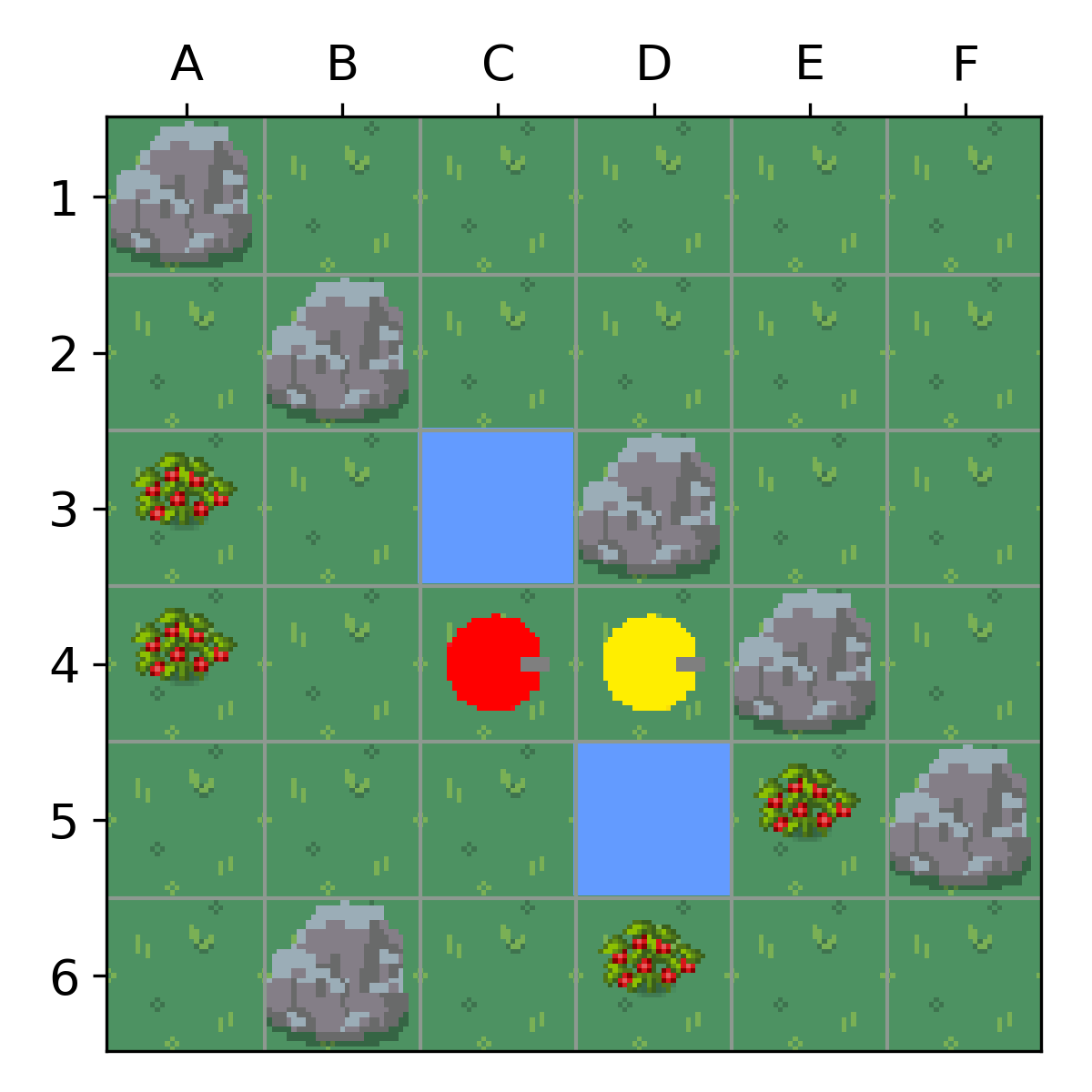}
            %\caption*{After 0.5}
            \centering \footnotesize Balanced, 0.5
            \label{fig:subsubfig2}
        \end{subfigure}
        \caption{Sample 1: A vs. B}
        \label{fig:sample1}
    \end{subfigure}%
    % Main subfigure 2
    \begin{subfigure}[t]{0.24\textwidth} % Half of total width
        \centering
        % Nested subfigure 2.1
        \begin{subfigure}[t]{0.5\textwidth} % 50% of the parent
            \includegraphics[width=\textwidth]{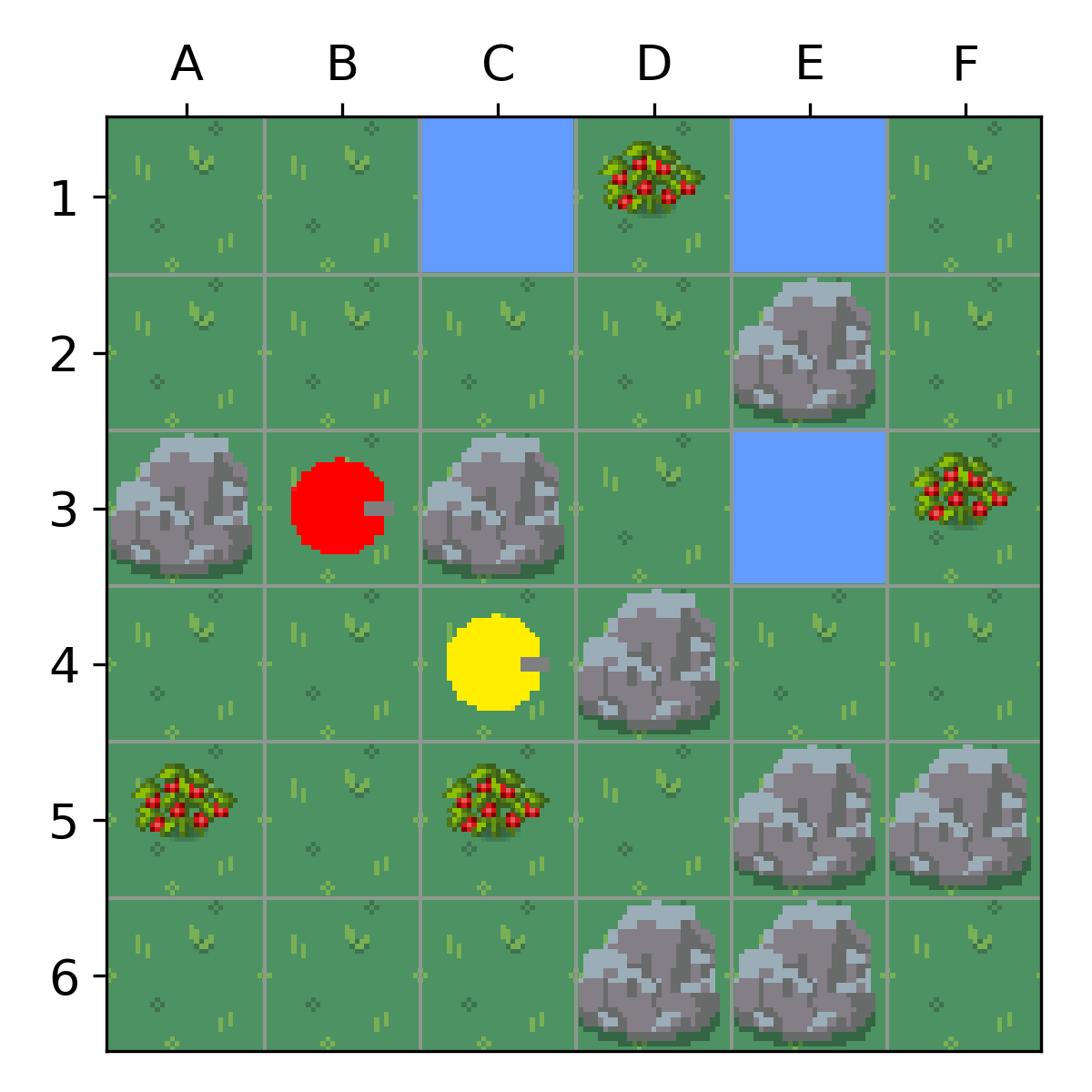}
            %\caption*{Before 0.0}
            \centering \footnotesize Unbalanced, 0.0
            \label{fig:subsubfig3}
        \end{subfigure}%
        %\hfill
        \begin{subfigure}[t]{0.5\textwidth} % 50% of the parent
            \centering
            \includegraphics[width=\textwidth]{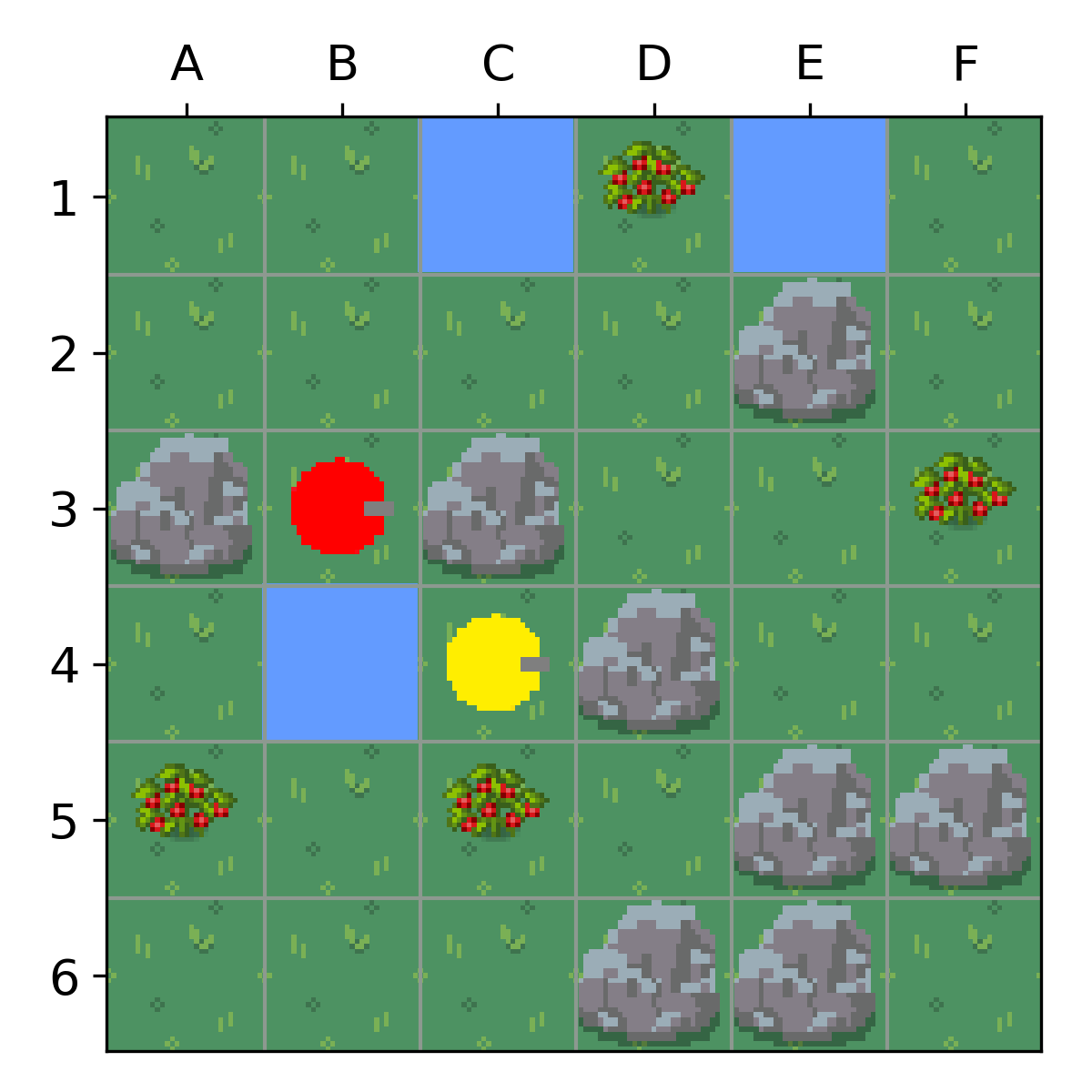}
            %\caption*{After 0.5}
            \centering \footnotesize Balanced, 0.5
            \label{fig:subsubfig4}
        \end{subfigure}
        \caption{Sample 2: A vs. C}
        \label{fig:sample2}
        % subfig 3
    \end{subfigure}
        \begin{subfigure}[t]{0.24\textwidth} % Half of total width
        \centering
        % Nested subfigure 1.1
        \begin{subfigure}[t]{0.5\textwidth} % 50% of the parent
            \centering
            \includegraphics[width=\textwidth]{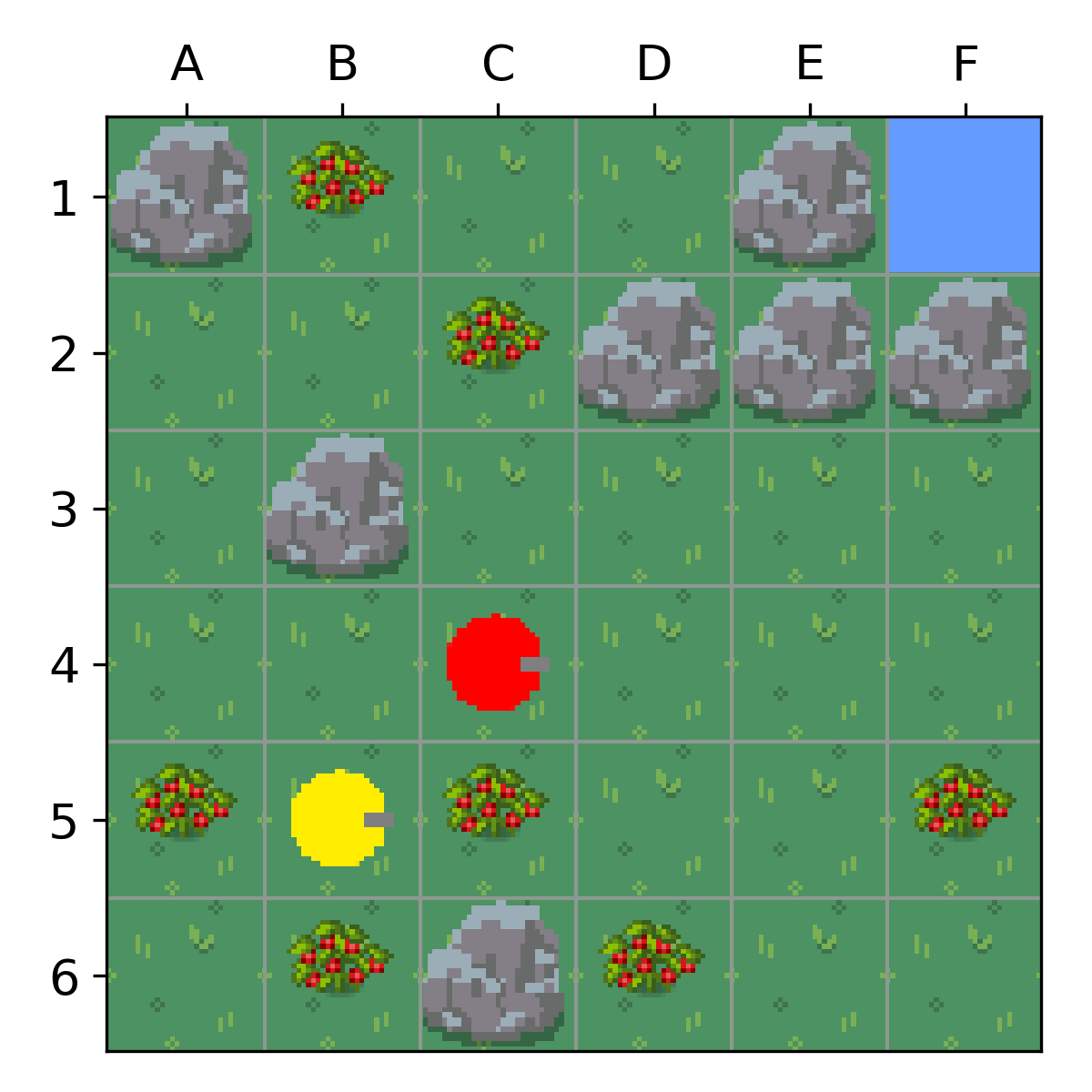}
            %\caption*{Before 1.0}
            \centering \footnotesize Unbalanced, 1.0
            \label{fig:subsubfig1}
        \end{subfigure}%
        \begin{subfigure}[t]{0.5\textwidth} % 50% of the parent
            \centering
            \includegraphics[width=\textwidth]{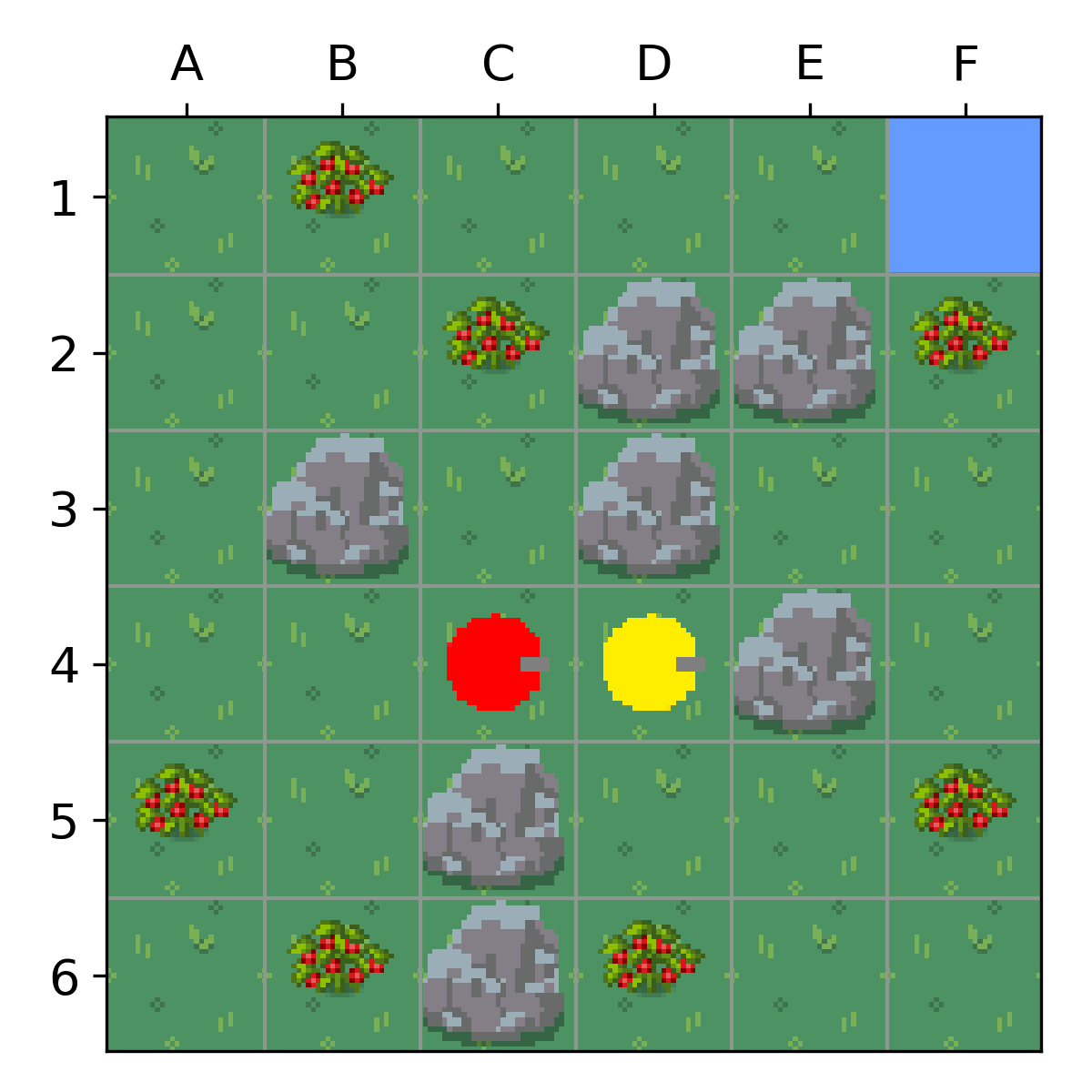}
            %\caption*{After 0.5}
            \centering \footnotesize Balanced, 0.5
            \label{fig:subsubfig2}
        \end{subfigure}
        \caption{Sample 3: A vs. D2}
        \label{fig:sample3}
    \end{subfigure}%
    % Main subfigure 4
    \begin{subfigure}[t]{0.24\textwidth} % Half of total width
        \centering
        % Nested subfigure 2.1
        \begin{subfigure}[t]{0.5\textwidth} % 50% of the parent
            \includegraphics[width=\textwidth]{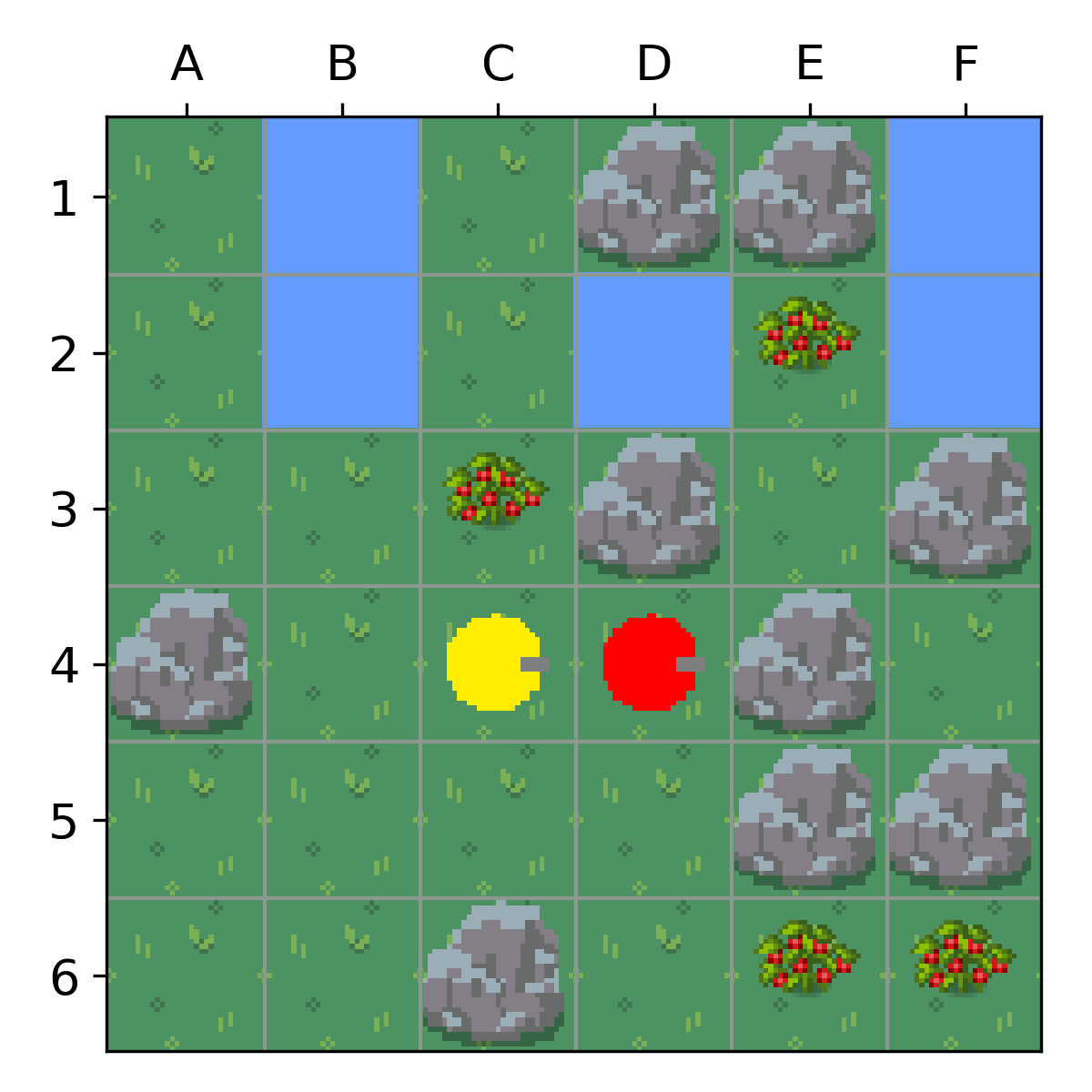}
            %\caption*{Before 0.0}
            \centering \footnotesize Unbalanced, 0.3
            \label{fig:subsubfig3}
        \end{subfigure}%
        %\hfill
        \begin{subfigure}[t]{0.5\textwidth} % 50% of the parent
            \centering
            \includegraphics[width=\textwidth]{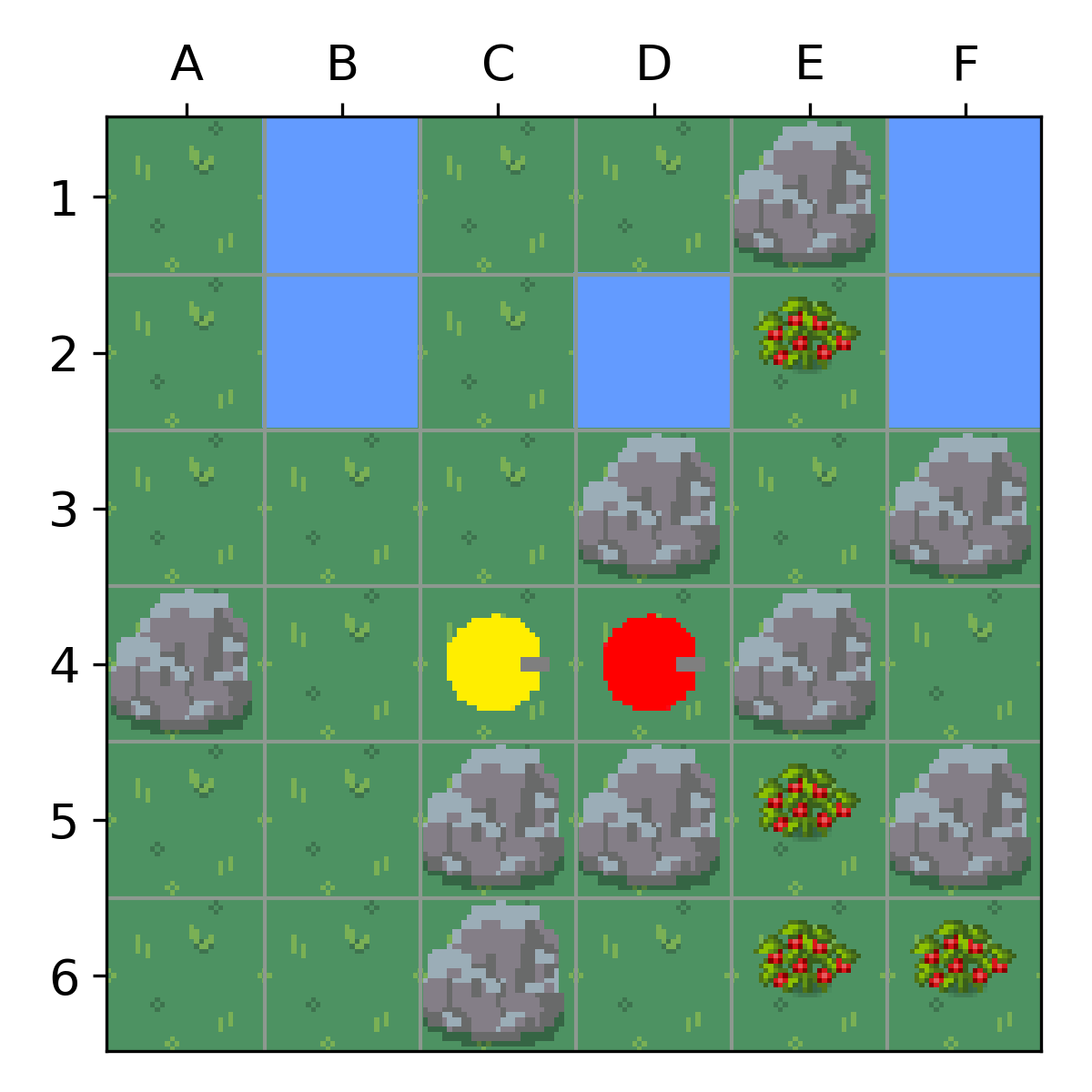}
            %\caption*{After 0.5}
            \centering \footnotesize Balanced, 0.5
            \label{fig:subsubfig4}
        \end{subfigure}
        \caption{Sample 4: A vs. D2}
        \label{fig:sample4}
    \end{subfigure}

    \caption{Generated samples from different models and archetype setups in comparison with the initial unbalanced version. A player of archetype A (red) is paired with different archetypes (yellow).}
    \label{fig:samples}
\end{figure}

\subsection{Generated Levels}
Figure~\ref{fig:samples} shows generated samples from different models and archetype setups. Sample 1 in Figure~\ref{fig:sample1} shows the setup of the \emph{Rock Agent} B (yellow) against a normal archetype (red).
The model achieved balance by swapping the tiles D3 and C4. Since the yellow agent can move over rock tiles, it can reach the area at the bottom right before the other player. If it could not move over rocks, this level would not be balanced.
Sample 2 shows the setup with the \emph{Handicap Agent} C (yellow), which can only move every second turn (Figure~\ref{fig:sample2}). The model achieves balance by separating the two players from each other by swapping B4 and E3 tiles. This prevents them from stealing food resources from each other, mitigating the yellow player's handicap.
The setup of the \emph{Food Agent} D2 (yellow), which already wins with three collected food resources instead of five, is shown in Figure~\ref{fig:sample3}. With several changes to the initial level, the model achieved balance by placing more resources on the side of the weaker agent (red). Also, the only water resource (tile A6) is now accessible to both players.

Sample 4 illustrates a limitation where the model achieves balance by excluding both players from access to food resources. This is a strategy that we see the model exploiting in certain cases to achieve a balanced state where both players technically win equally often - i.e., never~— but this outcome is not intended. This is possible due to the assumption in the reward function in the previous work~\cite{rupp_simulation_2024} that draws are always balanced, including when both players lose. 
In future work we thus aim to address this shortcoming by developing a solution that distinguishes between winnable and unwinnable draws.

\section{Conclusion}

We have proposed the use of reinforcement learning (RL) to balance tile-based game levels for asymmetric player archetypes entirely through the level design, extending a recently introduced method. Our results indicate that this approach can balance larger proportions of levels compared to a random search and a hill-climbing baseline. In addition, we optimize the action space of the original method, which accelerates training convergence.
Furthermore, we observe that the difficulty of learning balance increases with larger initial disparities in strength between the player archetypes, highlighting the challenge of dealing with highly unbalanced starting conditions.
A limitation of the RL approach is that it can exploit a strategy where neither player can win. While this technically ensures balance (both players win equally often), it is not the intended outcome. In future work we aim to address this shortcoming.

%% file: sample-sigconf-authordraft.bbl

\begin{thebibliography}{20}

%%% the Web address.

\ifx \showCODEN    \undefined \def \showCODEN     #1{\unskip}     \fi
\ifx \showDOI      \undefined \def \showDOI       #1{#1}\fi
\ifx \showISBNx    \undefined \def \showISBNx     #1{\unskip}     \fi
\ifx \showISBNxiii \undefined \def \showISBNxiii  #1{\unskip}     \fi
\ifx \showISSN     \undefined \def \showISSN      #1{\unskip}     \fi
\ifx \showLCCN     \undefined \def \showLCCN      #1{\unskip}     \fi
\ifx \shownote     \undefined \def \shownote      #1{#1}          \fi
\ifx \showarticletitle \undefined \def \showarticletitle #1{#1}   \fi
\ifx \showURL      \undefined \def \showURL       {\relax}        \fi
% invisible to TeX
\providecommand\bibfield[2]{#2}
\providecommand\bibinfo[2]{#2}
\providecommand\natexlab[1]{#1}
\providecommand\showeprint[2][]{arXiv:#2}

\bibitem[Beau and Bakkes(2016)]%
        {beau_automated_2016}
\bibfield{author}{\bibinfo{person}{Philipp Beau} {and} \bibinfo{person}{Sander Bakkes}.} \bibinfo{year}{2016}\natexlab{}.
\newblock \showarticletitle{Automated game balancing of asymmetric video games}. In \bibinfo{booktitle}{\emph{2016 {IEEE} {Conference} on {Computational} {Intelligence} and {Games} ({CIG})}}. \bibinfo{pages}{1--8}.
\newblock
\urldef\tempurl%
\url{https://doi.org/10.1109/CIG.2016.7860432}
\showDOI{\tempurl}


\bibitem[Becker and Görlich(2020)]%
        {becker_what_2020}
\bibfield{author}{\bibinfo{person}{Alexander Becker} {and} \bibinfo{person}{Daniel Görlich}.} \bibinfo{year}{2020}\natexlab{}.
\newblock \showarticletitle{What is {Game} {Balancing}? - {An} {Examination} of {Concepts}}.
\newblock \bibinfo{journal}{\emph{ParadigmPlus}}  \bibinfo{volume}{1} (\bibinfo{year}{2020}), \bibinfo{pages}{22--41}.
\newblock
\showISSN{2711-4627}
\urldef\tempurl%
\url{https://doi.org/10.55969/paradigmplus.v1n1a2}
\showDOI{\tempurl}


\bibitem[Earle et~al\mbox{.}(2021)]%
        {earle_learning_2021}
\bibfield{author}{\bibinfo{person}{Sam Earle}, \bibinfo{person}{Maria Edwards}, \bibinfo{person}{Ahmed Khalifa}, \bibinfo{person}{Philip Bontrager}, {and} \bibinfo{person}{Julian Togelius}.} \bibinfo{year}{2021}\natexlab{}.
\newblock \showarticletitle{{Learning Controllable Content Generators}}. In \bibinfo{booktitle}{\emph{2021 IEEE Conference on Games (CoG)}}. \bibinfo{pages}{1--9}.
\newblock
\urldef\tempurl%
\url{https://doi.org/10.1109/CoG52621.2021.9619159}
\showDOI{\tempurl}


\bibitem[Earle et~al\mbox{.}(2024)]%
        {earle_scaling_2024}
\bibfield{author}{\bibinfo{person}{Sam Earle}, \bibinfo{person}{Zehua Jiang}, {and} \bibinfo{person}{Julian Togelius}.} \bibinfo{year}{2024}\natexlab{}.
\newblock \showarticletitle{Scaling, {Control} and {Generalization} in {Reinforcement} {Learning} {Level} {Generators}}. In \bibinfo{booktitle}{\emph{2024 {IEEE} {Conference} on {Games} ({CoG})}}. \bibinfo{pages}{1--8}.
\newblock
\urldef\tempurl%
\url{https://doi.org/10.1109/CoG60054.2024.10645598}
\showDOI{\tempurl}


\bibitem[Jiang et~al\mbox{.}(2022)]%
        {jiang_learning_2022-4}
\bibfield{author}{\bibinfo{person}{Zehua Jiang}, \bibinfo{person}{Sam Earle}, \bibinfo{person}{Michael Green}, {and} \bibinfo{person}{Julian Togelius}.} \bibinfo{year}{2022}\natexlab{}.
\newblock \showarticletitle{Learning {Controllable} {3D} {Level} {Generators}}.
\newblock \bibinfo{journal}{\emph{Proceedings of the 17th International Conference on the Foundations of Digital Games}} (\bibinfo{year}{2022}).
\newblock
\urldef\tempurl%
\url{https://doi.org/10.1145/3555858.3563273}
\showDOI{\tempurl}


\bibitem[Khalifa et~al\mbox{.}(2020)]%
        {khalifa_pcgrl_2020}
\bibfield{author}{\bibinfo{person}{Ahmed Khalifa}, \bibinfo{person}{Philip Bontrager}, \bibinfo{person}{Sam Earle}, {and} \bibinfo{person}{Julian Togelius}.} \bibinfo{year}{2020}\natexlab{}.
\newblock \showarticletitle{PCGRL: {Procedural Content Generation via Reinforcement Learning}}. In \bibinfo{booktitle}{\emph{Proceedings of the {AAAI} {Conference} on {Artificial} {Intelligence} and {Interactive} {Digital} {Entertainment}}}, Vol.~\bibinfo{volume}{16}. \bibinfo{pages}{95--101}.
\newblock
\urldef\tempurl%
\url{https://doi.org/10.1609/aiide.v16i1.7416}
\showDOI{\tempurl}


\bibitem[Lanzi et~al\mbox{.}(2014)]%
        {lanzi_evolving_2014}
\bibfield{author}{\bibinfo{person}{Pier~Luca Lanzi}, \bibinfo{person}{Daniele Loiacono}, {and} \bibinfo{person}{Riccardo Stucchi}.} \bibinfo{year}{2014}\natexlab{}.
\newblock \showarticletitle{Evolving maps for match balancing in first person shooters}. In \bibinfo{booktitle}{\emph{2014 {IEEE} {Conference} on {Computational} {Intelligence} and {Games}}}.
\newblock
\urldef\tempurl%
\url{https://doi.org/10.1109/CIG.2014.6932901}
\showDOI{\tempurl}


\bibitem[Lara-Cabrera et~al\mbox{.}(2014)]%
        {lara-cabrera_balance_2014}
\bibfield{author}{\bibinfo{person}{Raúl Lara-Cabrera}, \bibinfo{person}{Carlos Cotta}, {and} \bibinfo{person}{Antonio~J. Fernández-Leiva}.} \bibinfo{year}{2014}\natexlab{}.
\newblock \showarticletitle{On balance and dynamism in procedural content generation with self-adaptive evolutionary algorithms}.
\newblock \bibinfo{journal}{\emph{Natural Computing}}  \bibinfo{volume}{13} (\bibinfo{year}{2014}), \bibinfo{pages}{157--168}.
\newblock
\showISSN{1572-9796}
\urldef\tempurl%
\url{https://doi.org/10.1007/s11047-014-9418-9}
\showDOI{\tempurl}


\bibitem[Mesentier~Silva et~al\mbox{.}(2019)]%
        {mesentier_silva_evolving_2019}
\bibfield{author}{\bibinfo{person}{Fernando~de Mesentier~Silva}, \bibinfo{person}{Rodrigo Canaan}, \bibinfo{person}{Scott Lee}, \bibinfo{person}{Matthew~C. Fontaine}, \bibinfo{person}{Julian Togelius}, {and} \bibinfo{person}{Amy~K. Hoover}.} \bibinfo{year}{2019}\natexlab{}.
\newblock \showarticletitle{Evolving the {Hearthstone} {Meta}}. In \bibinfo{booktitle}{\emph{2019 {IEEE} {Conf.} on {Games} ({CoG})}}.
\newblock
\urldef\tempurl%
\url{https://doi.org/10.1109/CIG.2019.8847966}
\showDOI{\tempurl}


\bibitem[Morosan and Poli(2017)]%
        {morosan_automated_2017}
\bibfield{author}{\bibinfo{person}{Mihail Morosan} {and} \bibinfo{person}{Riccardo Poli}.} \bibinfo{year}{2017}\natexlab{}.
\newblock \showarticletitle{Automated {Game} {Balancing} in {Ms} {PacMan} and {StarCraft} {Using} {Evolutionary} {Algorithms}}. In \bibinfo{booktitle}{\emph{Applications of {Evolutionary} {Computation}}} \emph{(\bibinfo{series}{Lecture {Notes} in {Computer} {Science}})}. \bibinfo{publisher}{Springer International Publishing}, \bibinfo{address}{Cham}, \bibinfo{pages}{377--392}.
\newblock
\showISBNx{978-3-319-55849-3}
\urldef\tempurl%
\url{https://doi.org/10.1007/978-3-319-55849-3_25}
\showDOI{\tempurl}


\bibitem[Pfau et~al\mbox{.}(2020)]%
        {pfau_dungeons_2020}
\bibfield{author}{\bibinfo{person}{Johannes Pfau}, \bibinfo{person}{Antonios Liapis}, \bibinfo{person}{Georg Volkmar}, \bibinfo{person}{Georgios~N. Yannakakis}, {and} \bibinfo{person}{Rainer Malaka}.} \bibinfo{year}{2020}\natexlab{}.
\newblock \showarticletitle{Dungeons \& {Replicants}: {Automated} {Game} {Balancing} via {Deep} {Player} {Behavior} {Modeling}}. In \bibinfo{booktitle}{\emph{{IEEE} {Conference} on {Games} ({CoG})}}. \bibinfo{pages}{431--438}.
\newblock
\urldef\tempurl%
\url{https://doi.org/10.1109/CoG47356.2020.9231958}
\showDOI{\tempurl}


\bibitem[Rupp et~al\mbox{.}(2023)]%
        {rupp_balancing_2023}
\bibfield{author}{\bibinfo{person}{Florian Rupp}, \bibinfo{person}{Manuel Eberhardinger}, {and} \bibinfo{person}{Kai Eckert}.} \bibinfo{year}{2023}\natexlab{}.
\newblock \showarticletitle{Balancing of competitive two-player {Game} {Levels} with {Reinforcement} {Learning}}. In \bibinfo{booktitle}{\emph{2023 {IEEE} {Conference} on {Games} ({CoG})}}.
\newblock
\urldef\tempurl%
\url{https://doi.org/10.1109/CoG57401.2023.10333248}
\showDOI{\tempurl}


\bibitem[Rupp et~al\mbox{.}(2024a)]%
        {rupp_simulation_2024}
\bibfield{author}{\bibinfo{person}{Florian Rupp}, \bibinfo{person}{Manuel Eberhardinger}, {and} \bibinfo{person}{Kai Eckert}.} \bibinfo{year}{2024}\natexlab{a}.
\newblock \showarticletitle{{Simulation-Driven Balancing of Competitive Game Levels With Reinforcement Learning}}.
\newblock \bibinfo{journal}{\emph{IEEE Trans. on Games}} \bibinfo{volume}{16}, \bibinfo{number}{4} (\bibinfo{year}{2024}), \bibinfo{pages}{903--913}.
\newblock
\urldef\tempurl%
\url{https://doi.org/10.1109/TG.2024.3399536}
\showDOI{\tempurl}


\bibitem[Rupp and Eckert(2024a)]%
        {rupp_gpcgrl_2024}
\bibfield{author}{\bibinfo{person}{Florian Rupp} {and} \bibinfo{person}{Kai Eckert}.} \bibinfo{year}{2024}\natexlab{a}.
\newblock \showarticletitle{G-PCGRL: Procedural Graph Data Generation via Reinforcement Learning}. In \bibinfo{booktitle}{\emph{2024 IEEE Conference on Games (CoG)}}. \bibinfo{pages}{1--8}.
\newblock
\urldef\tempurl%
\url{https://doi.org/10.1109/CoG60054.2024.10645633}
\showDOI{\tempurl}


\bibitem[Rupp and Eckert(2024b)]%
        {rupp_geevo_2024}
\bibfield{author}{\bibinfo{person}{Florian Rupp} {and} \bibinfo{person}{Kai Eckert}.} \bibinfo{year}{2024}\natexlab{b}.
\newblock \showarticletitle{{GEEvo}: {Game} {Economy} {Generation} and {Balancing} with {Evolutionary} {Algorithms}}. In \bibinfo{booktitle}{\emph{2024 {IEEE} {Congress} on {Evolutionary} {Computation} ({CEC})}}. \bibinfo{pages}{1--8}.
\newblock
\urldef\tempurl%
\url{https://doi.org/10.1109/CEC60901.2024.10612054}
\showDOI{\tempurl}


\bibitem[Rupp et~al\mbox{.}(2024b)]%
        {rupp_might_balanced_2024}
\bibfield{author}{\bibinfo{person}{Florian Rupp}, \bibinfo{person}{Alessandro Puddu}, \bibinfo{person}{Christian Becker-Asano}, {and} \bibinfo{person}{Kai Eckert}.} \bibinfo{year}{2024}\natexlab{b}.
\newblock \showarticletitle{{It might be balanced, but is it actually good? An Empirical Evaluation of Game Level Balancing}}. In \bibinfo{booktitle}{\emph{2024 IEEE Conference on Games (CoG)}}. \bibinfo{pages}{1--4}.
\newblock
\urldef\tempurl%
\url{https://doi.org/10.1109/CoG60054.2024.10645642}
\showDOI{\tempurl}


\bibitem[Schreiber and Romero(2021)]%
        {schreiber_game_2021}
\bibfield{author}{\bibinfo{person}{Ian Schreiber} {and} \bibinfo{person}{Brenda Romero}.} \bibinfo{year}{2021}\natexlab{}.
\newblock \bibinfo{booktitle}{\emph{Game {Balance}}}.
\newblock \bibinfo{publisher}{CRC Press}, \bibinfo{address}{Boca Raton}.
\newblock
\showISBNx{978-1-315-15642-2}
\urldef\tempurl%
\url{https://doi.org/10.1201/9781315156422}
\showDOI{\tempurl}


\bibitem[Schulman et~al\mbox{.}(2017)]%
        {schulman_proximal_2017}
\bibfield{author}{\bibinfo{person}{John Schulman}, \bibinfo{person}{Filip Wolski}, \bibinfo{person}{Prafulla Dhariwal}, \bibinfo{person}{Alec Radford}, {and} \bibinfo{person}{Oleg Klimov}.} \bibinfo{year}{2017}\natexlab{}.
\newblock \bibinfo{title}{Proximal {Policy} {Optimization} {Algorithms}}.
\newblock
\newblock
\urldef\tempurl%
\url{http://arxiv.org/abs/1707.06347}
\showURL{%
\tempurl}
\newblock
\shownote{arXiv:1707.06347}.


\bibitem[Suarez et~al\mbox{.}(2019)]%
        {suarez_neural_2019}
\bibfield{author}{\bibinfo{person}{Joseph Suarez}, \bibinfo{person}{Yilun Du}, \bibinfo{person}{Phillip Isola}, {and} \bibinfo{person}{Igor Mordatch}.} \bibinfo{year}{2019}\natexlab{}.
\newblock \bibinfo{title}{Neural {MMO}: {A} {Massively} {Multiagent} {Game} {Environment} for {Training} and {Evaluating} {Intelligent} {Agents}}.
\newblock
\newblock
\urldef\tempurl%
\url{http://arxiv.org/abs/1903.00784}
\showURL{%
\tempurl}
\newblock
\shownote{arXiv:1903.00784}.


\bibitem[Volz et~al\mbox{.}(2016)]%
        {volz_demonstrating_2016}
\bibfield{author}{\bibinfo{person}{Vanessa Volz}, \bibinfo{person}{Günter Rudolph}, {and} \bibinfo{person}{Boris Naujoks}.} \bibinfo{year}{2016}\natexlab{}.
\newblock \showarticletitle{Demonstrating the {Feasibility} of {Automatic} {Game} {Balancing}}. In \bibinfo{booktitle}{\emph{Proc. of the {Genetic} and {Evolutionary} {Computation} {Conf. (GECCO)}}}. \bibinfo{pages}{269--276}.
\newblock
\showISBNx{978-1-4503-4206-3}
\urldef\tempurl%
\url{https://doi.org/10.1145/2908812.2908913}
\showDOI{\tempurl}


\end{thebibliography}
